\documentclass[11pt]{article}
\usepackage[top=1in, bottom=1in, left=1in, right=1in]{geometry}
\usepackage{tgpagella}
\usepackage[utf8]{inputenc} %
\usepackage[T1]{fontenc}    %

\usepackage{url}
\usepackage[hidelinks]{hyperref}
\usepackage{amsmath,amsthm,amssymb,amsbsy}
\usepackage{paralist}
\usepackage{xcolor}
\usepackage{color}
\usepackage{graphicx}
\graphicspath{{./figs/}}
\usepackage{algorithm}
\usepackage{algorithmic}
\usepackage{comment}
\usepackage{multirow}

\usepackage{graphicx}
\usepackage{fancyhdr}
\usepackage[numbers,sort&compress]{natbib}
\usepackage{cleveref}
\usepackage{subcaption}

\theoremstyle{plain}

\newtheorem{definition}{Definition}
\theoremstyle{remark}
\newtheorem{remark}{Remark}

\newcommand{\e}{\begin{equation}}
\newcommand{\ee}{\end{equation}}
\newcommand{\en}{\begin{equation*}}
\newcommand{\een}{\end{equation*}}
\newcommand{\eqn}{\begin{eqnarray}}
\newcommand{\eeqn}{\end{eqnarray}}
\newcommand{\bmat}{\begin{bmatrix}}
\newcommand{\emat}{\end{bmatrix}}

\DeclareMathAlphabet\mathbfcal{OMS}{cmsy}{b}{n}

\newcommand{\dist}{\operatorname{dist}}

\def \st {\operatorname*{s.t.\ }}

\graphicspath{{./figs/}}

\newlength{\imgwidth}
\newboolean{twoColVersion}
\setboolean{twoColVersion}{false}
\newcommand{\twoCol}[2]{\ifthenelse{\boolean{twoColVersion}} {#1} {#2} }

\usepackage{tikz} 
\usetikzlibrary{positioning,shapes}
\usetikzlibrary{
    arrows.meta,
    positioning,
    calc,
    backgrounds
}
\usepackage{booktabs}
\usepackage{enumitem}
\DeclareRobustCommand{\rev}[1]{%
  \begingroup
  \color{blue}%
  \everymath\expandafter{\the\everymath\color{black}}%
  \everydisplay\expandafter{\the\everydisplay\color{black}}%
  #1%
  \endgroup
}

\newcommand{\Dact}{\mathcal{D}_{\mathrm{act}}}
\newcommand{\Ddyn}{\mathcal{D}_{\mathrm{dyn}}}
\newcommand{\Dpass}{\mathcal{D}_{\mathrm{pass}}}

\title{An Introduction to World Models for Generalist Agents}
\author{Zhihui Zhu\\[0.3em]
\small Department of Computer Science and Engineering\\
\small The Ohio State University\\
\small \href{mailto:zhu.3440@osu.edu}{\texttt{zhu.3440@osu.edu}}}
\date{October 20, 2026}
\begin{document}

\maketitle

 \begin{abstract}
World models have recently emerged as a promising
direction toward developing generalist agents
capable of understanding, predicting, and interacting
with diverse environments. However, despite growing
interest, the concept of a world model remains
loosely defined, encompassing approaches ranging
from predictive dynamics models and learned simulators
to models that jointly generate future observations
and actions. This introductory paper aims to clarify the concept
of world models and provide a unified perspective
on their role in developing generalist agents.
Starting from the partially observable Markov
decision process (POMDP) framework, we connect
world modeling to classical control, dynamical
systems, signal processing, and reinforcement
learning. We introduce different approaches to
world modeling, emphasizing the knowledge they
capture, their learning objectives, and how they
support decision-making. We further explore the
relationship between world modeling and policy
learning, highlighting how these capabilities
can be unified through conditional prediction
without requiring separate models.  Rather than offering an exhaustive survey of
existing methods, this paper focuses on fundamental
concepts, modeling principles, and connections,
with the goal of providing an accessible and
unified introduction to world models for
generalist agents.
\end{abstract}

"{\it The most incomprehensible thing about the universe is that it is comprehensible}"---Albert Einstein

\tableofcontents

\section{Introduction}

A long-standing goal of machine learning (ML) and artificial intelligence (AI) is to develop
\emph{generalist agents} that can perceive, reason, and act
effectively across diverse environments and tasks~\cite{reed2022generalist}.
Achieving this requires more than
recognizing patterns in observations or reproducing
actions from demonstrations. An agent must acquire
knowledge about how the world works, how it changes,
and how its actions influence future outcomes. Recently, \emph{world models} have emerged as a popular and promising direction toward this goal. Advances in generative modeling, video prediction,
robotics, and large language models (LLMs) have stimulated
growing interest in building agents that can learn
how the world works and use this knowledge to guide
their actions \cite{bruce2024genie,ding2025understanding}. This direction is particularly relevant to
physical intelligence, where agents must
understand and interact with the physical world
across diverse tasks and embodiments~\cite{black2026pi0visionlanguageactionflowmodel}.
The growing availability of images, videos,
and robotic interaction data provides new
opportunities to acquire such world knowledge
at scale~\cite{grauman2022ego4d,o2024open}.

Despite this growing interest, the meaning of
\emph{world model} remains surprisingly ambiguous.
The term has been used to describe a wide range of
models, including predictive dynamics models,
generative video models, learned simulators, and
models that jointly predict future observations and
actions. Even action policies can be viewed
as implicitly containing a model of the world.
These approaches differ substantially in their
learning objectives, representations, architectures,
and intended applications.  The idea of learning internal models of the
environment has a long history in AI and control. Model-based control
and reinforcement learning have traditionally
used models of system dynamics to predict the
consequences of actions and support planning \cite{sutton1991dyna,sutton1998reinforcement}.
The term \emph{world model} gained renewed
attention with a series of recent work \cite{ha2018recurrent,hafner2019learning,hafner2019dream,lecun2022path}. More recently, advances in video generation~\cite{liu2024sorareviewbackgroundtechnology}
and world-action models~\cite{ye2026world} have further blurred
the boundaries between predicting the world,
simulating interactions, and generating actions.

To clarify these concepts, it is useful to
consider the two words in \emph{world model}.
A \emph{model}, broadly speaking, is an
informative representation of an object or
system, constructed to describe, understand,
or predict its behavior \cite{wikipedia_model}.
A model need not reproduce every detail of
reality, but rather capture selected properties
relevant to its intended purpose. A mathematical model
of atmospheric dynamics, for example, captures
aspects of the physical system relevant to
weather forecasting. A related concept is \emph{simulation},
which involves constructing a model of a system
and conducting experiments with that model
to understand its behavior or evaluate
alternative strategies~\cite{shannon1975systems}.
The word \emph{world} highlights a broader
ambition: learning knowledge that is useful
across diverse environments and tasks, rather
than modeling a narrowly specified system
for a single application. For generalist agents,
such knowledge may include the structure of
environments, the evolution of their states,
the consequences of actions, and the relationships
between observations, actions, and outcomes.
Importantly, this does not require a world model
to capture every aspect of the environment.
Different models may represent different
components of the world, at different levels
of abstraction, depending on what information
is useful for prediction and decision-making.

\begin{figure*}[t]
\centering
\renewcommand{\bfdefault}{bx}
\begingroup
\definecolor{integratedNavy}{HTML}{19344D}
\definecolor{integratedBlue}{HTML}{3E73A6}
\definecolor{integratedTeal}{HTML}{287E7C}
\definecolor{integratedOrange}{HTML}{AE6822}
\definecolor{integratedPurple}{HTML}{77508F}
\resizebox{0.98\textwidth}{!}{%
\begin{tikzpicture}[
  >={Latex[length=2mm,width=1.4mm]},
  every node/.style={font=\small\sffamily,text=integratedNavy,align=center},
  data/.style={draw=integratedBlue!75,fill=integratedBlue!5,
    rounded corners=3pt,line width=.7pt,text width=6.2cm,
    minimum height=1.05cm,inner sep=5pt,font=\footnotesize\sffamily},
  model/.style={rounded corners=4pt,line width=.85pt,text width=4.35cm,
    minimum height=1.15cm,inner sep=7pt},
  application/.style={draw=integratedBlue!65,fill=integratedBlue!4,
    rounded corners=3pt,line width=.8pt,text width=3.7cm,
    minimum height=1.05cm,inner sep=6pt,font=\footnotesize\sffamily},
  flow/.style={->,draw=integratedNavy!65,line width=.85pt,rounded corners=3pt},
  support/.style={->,draw=integratedTeal,line width=.9pt,rounded corners=3pt},
  policyflow/.style={->,draw=integratedOrange,line width=.9pt,rounded corners=3pt}
]
\node[font=\normalsize\bfseries\sffamily] at (8.1,1.35)
  {Diverse experience for generalist agents};
\node[font=\footnotesize\sffamily,text=integratedBlue] at (3.9,.8)
  {Without action labels};
\node[font=\footnotesize\sffamily,text=integratedBlue] at (12.3,.8)
  {With action labels};
\node[data] (passive) at (3.9,0)
  {\textbf{Images, general videos,}\\[3pt]\textbf{egocentric human videos}};
\node[data] (actions) at (12.3,0)
  {\textbf{Demonstrations,}\\[3pt]\textbf{agent interactions}};
\draw[rounded corners=5pt,line width=.6pt,draw=integratedNavy!18,fill=integratedNavy!1]
  (.1,-1.35) rectangle (16.1,-4.25);
\node[font=\small\bfseries\sffamily] at (8.1,-1.73)
  {Reusable world knowledge for decisions};
\node[model,draw=integratedTeal,fill=integratedTeal!6] (wm) at (3.9,-3.03)
  {\textbf{World modeling $p_\theta$}};
\node[model,draw=integratedOrange,fill=integratedOrange!7] (policy) at (12.3,-3.03)
  {\textbf{Policy learning $\pi_\theta$}};
\draw[flow] (passive.south) -- (wm.north);
\draw[flow] (actions.south) -- (policy.north);
\draw[flow] (12.3,-1.0) -- (4.65,-1.0)
  -- ([xshift=.75cm]wm.north);
\draw[<->,draw=integratedPurple,line width=1pt]
  (wm.east) -- node[above=2pt,text width=3.0cm,minimum height=.65cm,
    inner sep=2pt,font=\footnotesize\sffamily,text=integratedPurple]
    {Shared knowledge\\and representations}
  node[below=2pt,text width=3.0cm,minimum height=.65cm,
    inner sep=2pt,font=\footnotesize\sffamily,text=integratedNavy!75]
    {Shared or separate\\networks} (policy.west);

\node[application] (plan) at (3.1,-6.0)
  {\textbf{Inference-time planning}};
\node[application,draw=integratedTeal,fill=integratedTeal!5,text width=2.7cm]
  (sim) at (7.3,-6.0)
  {\textbf{Learned simulator}};
\node[application,draw=integratedOrange,fill=integratedOrange!6] (improve) at (11.5,-6.0)
  {\textbf{Policy fine-tuning}};
\draw[support] (wm.south) -- (3.9,-4.5) -| (sim.north);
\draw[support] (sim.west) -- (plan.east);
\draw[support] (sim.east) -- node[above=18pt,font=\footnotesize\sffamily,
  text=integratedTeal,text width=1.6cm] {Imagined\\experience} (improve.west);
\draw[policyflow] ([xshift=-.1cm]policy.south)
  -- ([xshift=.7cm]improve.north);

\node[draw=integratedNavy!45,fill=integratedNavy!3,rounded corners=3pt,line width=.7pt,
  text width=15.6cm,minimum height=.75cm,inner sep=6pt,
  font=\small\bfseries\sffamily] (goal) at (8.1,-7.6)
  {Effective behavior across diverse environments and tasks};
\draw[support] (plan.south) -- (plan.south |- goal.north);
\draw[policyflow] (improve.south) -- (improve.south |- goal.north);
\draw[policyflow] ([xshift=1.8cm]policy.south)
  -- node[midway,right,inner sep=4pt,font=\footnotesize\sffamily,
    text=integratedOrange,text width=1.75cm] {Direct policy\\execution}
  ([xshift=6.0cm]goal.north);
\end{tikzpicture}%
}
\endgroup
\caption{\textbf{An integrated view of world modeling and policy learning.}
Using the physical world as an illustrative setting,
passive observations (e.g., images and videos) and
action-labeled trajectories (e.g., robotic interactions)
provide complementary learning signals for acquiring
world knowledge and learning action policies.
World modeling and policy learning are closely
connected and may share representations,
learning objectives, or model architectures.
The acquired world knowledge can support
inference-time planning, simulated experience
for policy improvement through reinforcement
learning, or direct action through a learned policy.
Together, these complementary approaches
support the development of generalist agents
capable of effective decision-making across
diverse environments, tasks, and embodiments.}
\label{fig:world-model-overview}
\end{figure*}
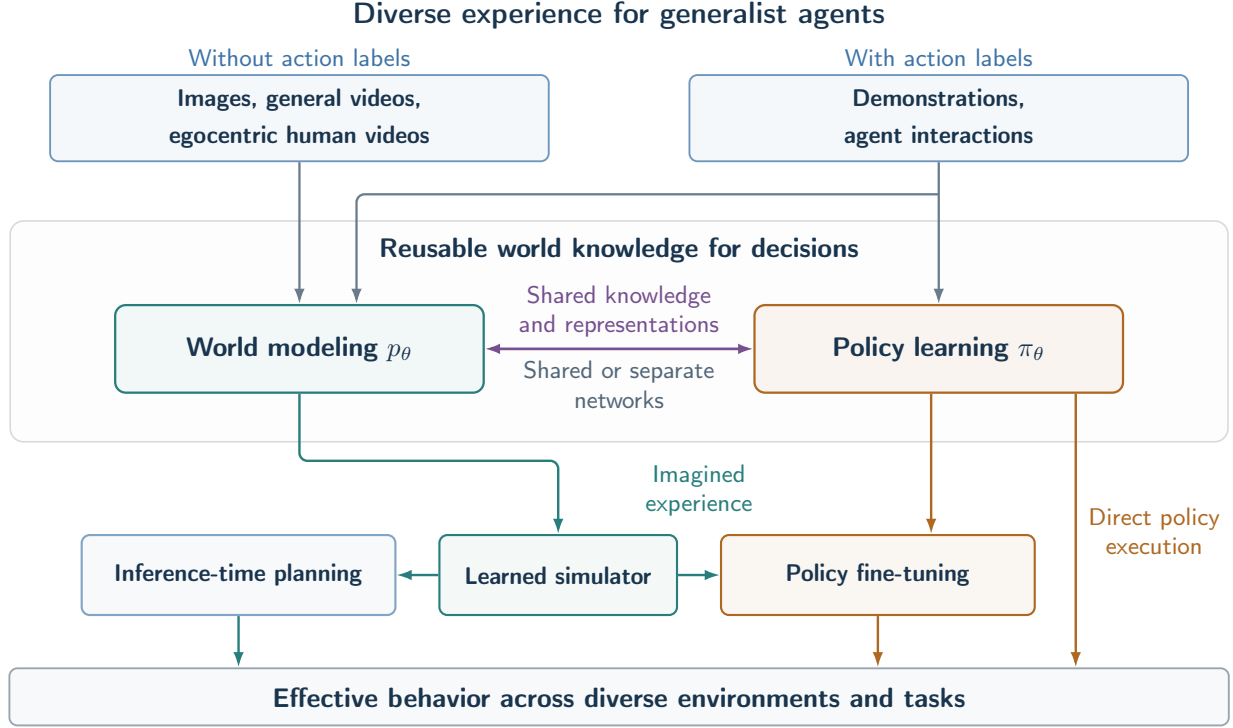

In this paper, we study world models from
the perspective of designing generalist agents,
whose objective is to maximize 
rewards across diverse environments and tasks.
We formalize this setting using partially observable Markov
decision processes (POMDPs)~\cite{sutton1998reinforcement},
which provide a unified framework for describing
environment states, observations, actions, and
dynamics.
We adopt a broad definition of a
world model as a learned representation
that captures aspects of an environment's
structure, state, or dynamics to support
understanding, prediction, or decision-making.
Under this perspective, different approaches
to world modeling can be characterized by
the environmental knowledge they capture
and the relationships they learn. For example, observation-space world models
predict future observations, while latent-space
models learn representations and their evolution.
Passive predictive models learn temporal
regularities from observations
~\cite{gao2022simvp}, whereas
action-conditioned models predict how the
environment responds to actions
~\cite{hafner2019learning}.

A policy, in contrast, selects actions based
on the available information and task.
Although policy learning does not necessarily
involve explicit world modeling, effective
decision-making often requires implicitly
capturing relevant aspects of the environment
and its dynamics.
Thus, world modeling and policy learning
represent distinct but closely related
capabilities, both \emph{relying on knowledge
of the world to varying degrees to support effective behavior}. As illustrated in \Cref{fig:world-model-overview},
they can leverage diverse data sources and
shared representations, while learned world
models can further support inference-time
planning and policy improvement through
simulated experience. More fundamentally, world modeling and policy
learning can be connected through a unified
probabilistic perspective.
A policy predicts which action to take given
the current context, a forward dynamics model
predicts the consequences of an action, and
an inverse dynamics model predicts the action
associated with an observed transition.
These can be viewed as different conditional
distributions derived from a joint distribution
over actions and future observations.
This perspective motivates recent
\emph{world-action models}, which integrate
world prediction and action generation through
shared representations, joint learning objectives,
or interacting generative architectures~\cite{wu2024unleashing,ye2026world}.

LLMs provide another
intriguing perspective. Trained on internet-scale
text through next-token prediction, followed
by fine-tuning and reinforcement learning,
LLMs have demonstrated broad capabilities
in reasoning, tool use, and interactive
problem-solving~\cite{yao2023react}. Although next-token prediction
is not explicitly designed for learning
environment dynamics, the training data
contains extensive descriptions of events,
actions, interactions, and their consequences.
Depending on the conditioning context and
prediction target, the same autoregressive
objective can implicitly provide supervision
for policies, forward dynamics, inverse
dynamics, and other forms of prediction.
This suggests that general-purpose
predictive learning may acquire implicit
representations of the world
~\cite{li2022emergent}, which can be further
strengthened through interaction-based
training~\cite{chae2025web,gu2025webdreamer,zhang2026earlyexperience}.

Our goal in this paper is to provide a unified
introduction to world models from the perspective
of learning generalist agents. Rather than focusing
on specific architectures or implementation details,
we emphasize the underlying modeling principles,
aiming to clarify the relationships among different
approaches to world modeling, their learning
objectives, and their roles in supporting
decision-making. Inspired by the success of
large-scale predictive learning in LLMs,
we also highlight how these principles may
enable generalist agents to leverage diverse
data sources, ranging from passive videos
and egocentric human observations to
action-labeled demonstrations and interaction
trajectories. By extracting complementary
learning signals from such data, world modeling
and policy learning may jointly support the
acquisition of transferable knowledge across
environments, tasks, and embodiments.
We do not attempt an exhaustive survey of
the rapidly growing literature, for which
we refer readers to several recent
surveys~\cite{ding2025understanding,hou2026world}.
Instead, we focus on representative approaches
that illustrate the key concepts and connections.

The remainder of this paper is organized as follows.
We first formulate the generalist-agent problem
using the POMDP framework in
\Cref{sec:generalist-agents-formulation}.
Following the perspective of~\cite{recht2019tour},
we then draw connections to classical control,
dynamical systems, signal processing, and
reinforcement learning in \Cref{sec:control-asp}.
In \Cref{sec:models-methods}, we introduce
policy learning and different approaches to
world modeling, highlighting their learning
objectives, representations, and roles in
supporting decision-making. Finally, we conclude in \Cref{sec:discussion-conclusion},
we conclude with a discussion of some other key challenges
and future directions, including generalization
across embodiments, scaling data and learning
signals, state and action representations,
high-level reasoning, and the evaluation of
world models for decision-making.

\section{Generalist Agents: A Sequential Decision-Making Perspective}
\label{sec:generalist-agents}

\subsection{General Formulation: partially observable Markov decision processe}
\label{sec:generalist-agents-formulation}

A \emph{generalist agent} can perform diverse tasks across environments, adapting its behavior to new situations under partial observability. A central research question is how agents can acquire transferable knowledge from heterogeneous data and interaction experience to support effective decision-making and generalization. Our research investigates the learning principles that enable agents to leverage both passive observations and active interactions to understand their environments, plan actions, and adapt to unfamiliar tasks~\cite{reed2022generalist,hafner2025mastering}.

\begin{table}[tbp]
\centering
\caption{Notation used throughout the paper.}
\label{tab:notation}
\small
\begin{tabular}{p{0.27\textwidth}p{0.65\textwidth}}
\toprule
Notation & Meaning \\
\midrule
$s_t,o_t,a_t,h_t$ & State, observation, action, and interaction history. \\
$e,g$ & Environment and task. \\
$\pi,\pi_\theta$ & Generic action policy and parameterized policy. \\
$p,p_\theta$ & Underlying  distribution, and a learned probabilistic model. \\
$\phi, z_t = \phi(h_t)$ & Latent encoder and latent representation. \\
$\theta$ & Generic model parameters, which may vary across models. \\
\bottomrule
\end{tabular}
\end{table}

To conceptualize interaction in machine learning, we adopt a general framework that encompasses a wide range of settings, including reinforcement learning, control, filtering, and system identification~\cite{kaelbling1998planning,sutton1998reinforcement}. This framework emphasizes the sequential nature of interaction. To capture the generality of intelligent agents, we explicitly model the diversity of environments and tasks they may encounter, rather than restricting attention to a single fixed setting.
Within a world, such as the physical world, the digital world, or a scientific domain, an agent may encounter many environments, denoted by $
e$,
which share common structure but differ in their specific configurations. For example, in the physical world, environments may correspond to different homes, restaurants, supermarkets, coffee shops, or offices.  Within a given environment $e$, the agent is asked to perform a task $g$, which specifies a goal or desired outcome. Table~\ref{tab:notation} summarizes the recurring symbols.

\begin{remark}
For simplicity, we assume a fixed agent embodiment throughout
the discussion, including its observation and action spaces.
In practice, generalist agents may operate across different
embodiments, such as robotic platforms with different sensors
and actuators or digital agents with different interfaces and
available tools. The framework can be extended to accommodate
such diversity by introducing an embodiment variable $b$,
with the observation and action spaces, as well as the
corresponding dynamics and policies, conditioned on $b$.
Ideally, a generalist agent should learn transferable
knowledge and capabilities that generalize not only across
environments and tasks, but also across embodiments.
\end{remark}

\begin{figure}[tbp]
\centering
\begin{tikzpicture}[
  >=stealth,
  every node/.style={font=\small},
  flow/.style={->,line width=1.2pt,draw=black!80},
  icon/.style={line width=1pt,draw=black!85}
]
\begin{scope}[shift={(-3.6,0.7)}]
  \draw[icon,fill=blue!4,rounded corners=9pt] (-0.65,-0.75) rectangle (0.65,0.75);
  \foreach \y in {-0.42,0,0.42} {
    \foreach \z in {-0.28,0.28} {
      \draw[draw=blue!55!black,line width=0.65pt] (-0.36,\y) -- (0,\z);
      \draw[draw=blue!55!black,line width=0.65pt] (0,\z) -- (0.36,\y);
    }
  }
  \foreach \y in {-0.42,0,0.42} {
    \fill[blue!65!black] (-0.36,\y) circle (2pt);
    \fill[blue!65!black] (0.36,\y) circle (2pt);
  }
  \foreach \y in {-0.28,0.28} {\fill[blue!65!black] (0,\y) circle (2pt);}
\end{scope}
\node[font=\large,anchor=east] at (-4.45,0.7) {Agent};
\node[align=center,anchor=north] at (-3.6,-0.3)
  {History $h_t$, task $g$\\[3pt]$h_{t+1}=(h_t,a_t,o_{t+1})$};
\begin{scope}[shift={(3.6,0.7)}]
  \draw[icon,fill=black!3] (0,0) circle (0.65);
  \draw[icon] (0,0) ellipse (0.27 and 0.65);
  \draw[icon] (-0.65,0) -- (0.65,0);
  \draw[icon] (-0.56,0.33) .. controls (-0.2,0.18) and (0.2,0.18) .. (0.56,0.33);
  \draw[icon] (-0.56,-0.33) .. controls (-0.2,-0.18) and (0.2,-0.18) .. (0.56,-0.33);
\end{scope}
\node[font=\large,anchor=west] at (4.45,0.7) {Environment};
\node at (3.6,-0.52) {state $s_t$};
\begin{scope}[shift={(0,-2.15)}]
  \draw[icon,fill=teal!5] (-0.75,0)
    .. controls (-0.32,0.63) and (0.32,0.63) .. (0.75,0)
    .. controls (0.32,-0.63) and (-0.32,-0.63) .. cycle;
  \draw[icon,fill=white] (0,0) circle (0.27);
  \fill[black!85] (0,0) circle (0.12);
\end{scope}
\node[font=\large] at (0,-2.95) {Observation};
\draw[flow] (-3.6,1.58) .. controls (-3.6,3.2) and (3.6,3.2) .. (3.6,1.48);
\node[align=center] at (0,3.45) {Action\\[3pt]$a_t\sim\pi(\cdot\mid h_t,g)$};
\draw[flow] (3.6,-0.85) .. controls (3.6,-2.15) and (2.1,-2.15) .. (0.92,-2.15);
\node[align=center] at (4.0,-2.7) {State transition\\[3pt]$s_{t+1}\sim p(\cdot\mid s_t,a_t,e)$};
\draw[flow] (-0.92,-2.15) .. controls (-2.1,-2.15) and (-3.6,-2.15) .. (-3.6,-1.5);
\node[align=center] at (-4.0,-2.7) {Partial observation\\[3pt]$o_{t+1}\sim p(\cdot\mid s_{t+1},e)$};
\end{tikzpicture}
\caption{The POMDP interaction loop for generalist agents
across diverse environments and tasks.
Given its interaction history and task, the agent
selects an action. The state evolves according
to $p(s_{t+1}\mid s_t,a_t,e)$, and the environment generates
an observation through $p(o_{t+1}\mid s_{t+1},e)$.
The agent incorporates the new observation into its
history before acting again. The environment $e$
and task remain fixed within an episode but may
vary across episodes, requiring the agent to adapt
to different environments and objectives.}
\label{fig:pomdp-interaction}
\end{figure}
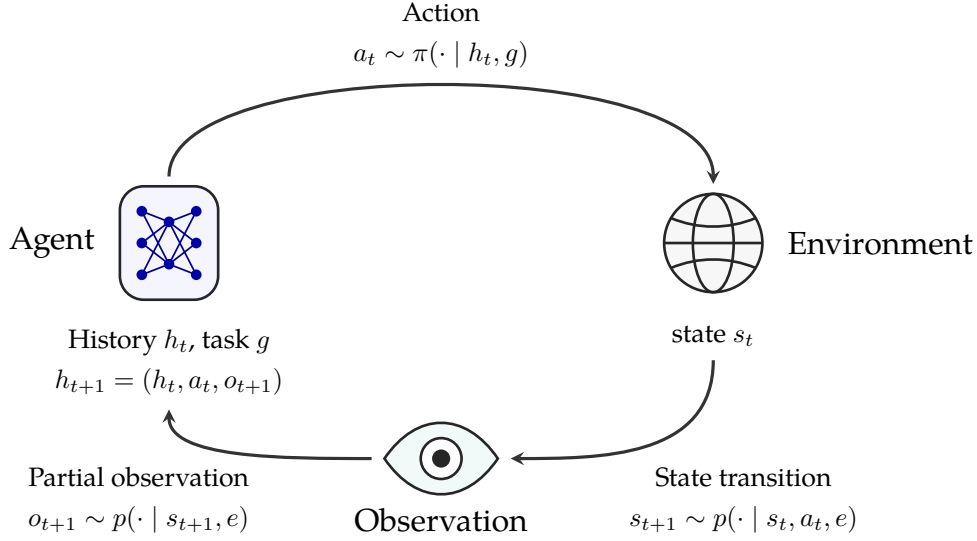

As illustrated in Figure~\ref{fig:pomdp-interaction}, the POMDP provides a principled mathematical framework for modeling sequential interaction and decision-making. Adapting the notations of state-action pairs $(s,a)$, 
each environment $e$ has a state space $\mathcal S_{e}$, observation space $\mathcal O_{e}$, and action space $\mathcal A_{e}$. Within one episode the environment remains fixed, while the state evolves according to its transition dynamics. At time $t$, the world has a possibly underlying state $s_t\in\mathcal S_{e}$, the agent receives an observation $o_t\in\mathcal O_{e}$, and the agent executes an action $a_t\in\mathcal A_{e}$. The state evolves according to
\begin{equation}
s_{t+1}
\sim
p(\cdot \mid s_t,a_t,e),
\end{equation}
while the agent may receive only a potentially partial observation of the underlying state,
\begin{equation}
o_t
\sim
p(\cdot \mid s_t,e).
\end{equation}
Here, for notational simplicity, we simply use $p$
for both the transition and observation
distributions, distinguished by their
predicted variables and conditioning
information. These correspond to the
transition and observation models,
often denoted by $P$ and $\Omega$,
respectively, in standard POMDP formulations.
An episode begins with an initial state
$s_0\sim p(\cdot\mid e)$ and observation
$o_0\sim p(\cdot\mid s_0,e)$.
We assume that the environment and task remain
fixed within an episode, and that the transition
and observation mechanisms are time-invariant
within each environment, although they may
differ across environments.
Importantly, conditioning on $e$ characterizes
the dependence of the underlying distributions
on the environment, but does not imply that
the environment identity is known to the agent.
We assume that the task $g$ is provided,
while the agent must infer relevant information
about the environment from its observations
and interaction history.

The interaction history up to time $t$ consists of the sequence of observations and actions:
\begin{equation}
h_t = (o_0, a_0, \ldots, a_{t-1}, o_t).
\end{equation}
The history update is $h_{t+1}=(h_t,a_t,o_{t+1})$. We assume a horizon of $T$ transitions. At time $t$, the agent selects an action based on its interaction history $h_t$, together with the task $g$. This decision-making process is characterized by a \emph{policy}.

\begin{definition}[Policy]
A \emph{policy} is a mapping from the interaction history to a distribution over actions:
\begin{equation}
a_t \sim \pi(\cdot \mid h_t,g).
\end{equation}
\end{definition}
Here, we adopt the standard stochastic policy formulation, where $\pi(a_t \mid h_t,g)$ specifies the probability (or probability density) of selecting action $a_t$. A deterministic policy is a special case that assigns all probability mass to a single action. Explicit conditioning on $g$ allows a single policy to support multiple tasks. {For a fixed, known environment with fully observed Markov state, a policy can depend on $s_t$ and $g$ rather than the full history under standard Markov decision process assumptions. A finite-horizon policy may also depend on time. If the environment is unknown, history can remain useful even when the physical state is observed, because past transitions help identify the environment dynamics.} In words, the policy maps the agent's available information and task objective to actions, making it a central object of learning and optimization in reinforcement learning and control. From a predictive modeling perspective, a policy can also be viewed as an \emph{action model} that models the conditional distribution of actions given the interaction history and task. We revisit this perspective when discussing different learning approaches.

A central challenge for generalist agents is
to operate effectively across diverse
environments and tasks.
This challenge is particularly pronounced
in the physical world, where agents may
encounter different environments, such as
kitchens, bedrooms, offices, warehouses,
and laboratories, and perform a wide range
of tasks, such as navigation, object
manipulation, and tool use.
Different environments involve variations
in objects, layouts, physical conditions,
and interaction dynamics, while different
tasks impose distinct goals and constraints.
Rather than learning a separate model for
every environment and task, a generalist
agent should acquire reusable world knowledge
about objects, physical dynamics, interactions,
and action consequences, and adapt this
knowledge to the current environment and task. 

Thus, the objective is to learn a generalist agent capable of performing diverse tasks across environments. Formally, let $r_g(s_t,a_t)$ denote the task-specific reward function, which measures progress toward the objective of task $g$. We seek\footnote{For simplicity, we assume that the reward function $r_g$ is given and omit the discount factor. In practice, designing an appropriate reward function is itself an important engineering problem. For example, a robotic task may use task success, execution time, energy consumption, or safety as part of the reward.}
\begin{equation}
\begin{aligned}
\max_{\pi}\quad
&
\mathbb{E}_{e,g}
\left[
\mathbb{E}_{s_t,o_t,a_t}\!\left[
\sum_{t=0}^{T-1}
r_{g}(s_t,a_t)
\right]
\right] \\
\text{s.t.}\quad
&
a_t\sim\pi(\cdot\mid h_t,g), \quad s_{t+1}\sim p(\cdot\mid s_t,a_t,e), \quad o_t
\sim
p(\cdot \mid s_t,e).
\end{aligned}
\label{eq:max-reward-obj}\end{equation}
The outer expectation is taken over
environments $e$ and tasks $g$,
while the inner expectation accounts for
the stochasticity of the initial state,
observations, state transitions, and
policy actions.
For simplicity, we consider a finite-horizon
objective with action-dependent rewards
and no additional terminal reward.
The formulation can be readily extended
to accommodate other reward structures,
including terminal rewards and discounted
cumulative rewards.
The objective is to learn a single policy
that achieves high expected reward across
diverse environments and tasks, rather
than specializing to a single fixed setting.

The above formulation is closely related to classical reinforcement learning, particularly its multi-task and partially observable variants \cite{kaelbling1998planning,sutton1998reinforcement,taylor2009transfer}. However, learning generalist agents introduces additional challenges beyond optimizing a policy in a fixed environment. First, agents must generalize across diverse and potentially unseen environments and tasks, rather than specialize in a single setting. Second, learning effective behaviors from scratch through trial and error may be prohibitively expensive, motivating the acquisition of transferable knowledge from heterogeneous data and prior experience. Third, unlike many standard RL benchmarks, real-world environments may not provide inexpensive, resettable interactions or accurate simulators, making exploration and policy improvement substantially more challenging. These considerations shift the focus from policy optimization alone toward understanding what knowledge agents should acquire, how they should learn it from diverse sources, and how they can use it to adapt and improve with limited interaction.

These challenges motivate approaches that
go beyond learning action policies from
trial-and-error interactions.
In particular, generalist agents may benefit
from learning predictive world models and
reusable representations, which can support
policy learning, planning, and adaptation
across environments and tasks. In the next section, we will discuss these approaches through the lens of what is learned, what learning signals are available, and how learned knowledge supports decision-making. Before doing so, we first connect our formulation to established fields of sequential decision-making, highlighting their underlying assumptions and objectives to provide a foundation for understanding the learning approaches discussed later.

\subsection{Sequential Decision-Making: Control, Estimation, and Learning}
\label{sec:control-asp}

The formulation \eqref{eq:max-reward-obj} connects to a rich body of work on dynamical systems, control, filtering, system identification, and reinforcement learning. These fields study different aspects of sequential interaction and can be understood within a common framework by considering two fundamental questions: whether the underlying state is directly observable and whether the environment dynamics are known. These distinctions help clarify the relationships among classical approaches and their relevance to learning generalist agents.

{In classical control and dynamical signal processing, the system is often written as}~\cite{kalman1960new,anderson2005optimal,aastrom2008feedback}
\begin{equation}
    s_{t+1} = f(s_t,a_t,w_t), \qquad
    o_t = q(s_t,v_t),
\label{eq:dynamics}\end{equation}
where $w_t$ and $v_t$ denote process and measurement noise. {In this subsection, we consider a fixed environment and task, suppressing $e$ and $g$ from the notation. The functions $f$ and $q$, together with the noise distributions, induce the transition and observation conditionals denoted by $p$.} In this context, a policy plays the role of a controller that selects actions based on the available information. {Classical control may often use deterministic controllers. With a slight abuse of notation, we write $a_t=\pi(h_t)$ for a policy that assigns all probability mass to one action. A finite-horizon controller can depend on time, which is already encoded by the length of $h_t$.}

\subsubsection{Known Dynamics and Fully Observed State: Control and Dynamic Programming.}
When the state $s_t$ is fully observed and the transition model $f(s_t,a_t,w_t)$ is known, the problem reduces to classical sequential decision making and optimal control. Let $\xi(s_t,a_t)$ denote the instantaneous cost incurred at time $t$ when taking action $a_t$ in state $s_t$.
Following the standard formulation in optimal control, we consider a finite-horizon problem with horizon $T$, where the objective is to minimize the expected cumulative cost:
\begin{equation}
    \begin{split}
    & \min_{\pi} \; \mathbb{E}_{w_{0:T-1}}\left[\sum_{t=0}^{T-1} \xi(s_t,a_t)\right],\\
    & \st  s_{t+1} = f(s_t,a_t,w_t), \quad a_t = \pi(h_t).
    \end{split}
    \label{eq:control-objective}
\end{equation}
or equivalently as reward maximization with reward $-\xi(s_t,a_t)$. When both the system dynamics and the objective function are known, a rich collection of methods has been developed to solve \eqref{eq:control-objective}. For instance, dynamic programming exploits Bellman’s principle of optimality to decompose the global optimization problem into a sequence of local optimization problems, yielding an optimal policy under suitable assumptions \cite{bertsekas1995dynamic}.

A classical example is the \emph{linear quadratic regulator} (LQR):
\begin{equation}
    \begin{split}
    & \min_{\pi} \; \mathbb{E}_{w_{0:T-1}}\left[\sum_{t=0}^{T-1} \left(s_t^\top Q s_t + a_t^\top R a_t\right) \right], \\
    & s_{t+1} = A s_t + B a_t + w_t, \qquad a_t=\pi_t(s_t),
    \end{split}
\label{eq:lqr-objective}\end{equation}
which considers linear dynamics and quadratic costs. Here, $Q\succeq 0$ and $R\succ 0$ penalize state deviations and control
effort, respectively. Under standard assumptions, the optimal controller
is a linear state-feedback law, $a_t=-K_t s_t$, where the gain $K_t$
is obtained from the Riccati recursion~\cite{anderson2007optimal}.

\emph{Model predictive control} (MPC) extends this optimization-based
perspective by repeatedly solving a finite-horizon control problem
using the current state or state estimate. At each time step, only
the first planned action is executed, after which the optimization
is repeated using updated information~\cite{rawlings2017model}.
Unlike dynamic programming, which characterizes an optimal policy
through a value function, MPC computes actions through repeated
online optimization. Together, these approaches illustrate the
known-dynamics setting, where the primary challenge is planning
and selecting actions rather than learning the environment dynamics.

\subsubsection{Known Dynamics and Partially Observed State: Filtering and State Estimation.}
{When the transition and observation mechanisms in \eqref{eq:dynamics} are known but $s_t$ is not directly observed, the agent must infer the state from its history $h_t=(o_0,a_0,\ldots,a_{t-1},o_t)$. This is the classical problem of \emph{filtering} or \emph{state estimation}}~\cite{kalman1960new,anderson2007optimal}.

A general formulation of the problem is to mantain the posterior distribution (or \emph{belief state}) over states given the history, $p(s_t\mid h_t)$. This posterior summarizes everything known about the state. Noting that the history $h_t$ can be decomposed as $h_t = (h_{t-1}, a_{t-1}, o_t)$, we first compute the prior distribution over states at time $t$ given the previous history and action using the transition model:
\begin{equation}
    \begin{split}
    p(s_t \mid h_{t-1}, a_{t-1}) &= \int p(s_t , s_{t-1} \mid h_{t-1}, a_{t-1})  ds_{t-1},\\
    & = \int p(s_t \mid s_{t-1}, a_{t-1}) p(s_{t-1} \mid h_{t-1}) ds_{t-1}, \quad \text{(prediction step)}
\end{split}
\end{equation}
where the last equlity follows from the Markov property of the state transition and the fact that since $a_{t-1}$ does not alter the posterior distribution of $s_{t-1}$ after the past observations. We then incorporate the new observation $o_t$ through Bayes’ rule:
\begin{equation}
    \begin{split}
    p(s_t \mid h_t) = \frac{p(o_t \mid s_t, h_{t-1}, a_{t-1}) p(s_t \mid h_{t-1}, a_{t-1})}{p(o_t \mid h_{t-1}, a_{t-1})}  \propto p(o_t \mid s_t) p(s_t \mid h_{t-1}, a_{t-1}). \quad \text{(update step)}
    \end{split}
\end{equation}
Thus, in principle, the belief state $p(s_t \mid h_t)$ can be computed recursively from the previous belief $p(s_{t-1} \mid h_{t-1})$, the transition model $p(s_{t} \mid s_{t-1}, a_{t-1})$, and the observation model $p(o_t \mid s_t)$, which are induced by the dynamics and observation models in \eqref{eq:dynamics}. If a point estimate is desired, one may compute the posterior mean or the posterior mode depends on the application. 

This recursive Bayesian formulation underlies a broad family of
filtering algorithms. In practice, the belief state is often represented in a parametric form (e.g., Gaussian) or approximated by a finite set of particles, leading to algorithms such as Kalman filtering and particle filtering. For instance, Kalman filtering is the classical solution for linear-Gaussian systems,
\begin{equation}
    s_{t+1}=A s_t + B a_t + w_t, \qquad
    o_t=C s_t + v_t,
\end{equation}
where $w_t$ and $v_t$ are independent Gaussian noise processes. {With a Gaussian initial-state distribution, the posterior $p(s_t\mid h_t)$ remains Gaussian for all $t$, and both the prediction and update steps admit closed-form solutions.}

Filtering and control address complementary aspects of sequential
interaction: filtering estimates the underlying state, while control
selects actions to achieve an objective. In the classical linear-quadratic-Gaussian setting, the separation
principle establishes that the optimal controller can be obtained
by combining a Kalman filter with the LQR introduced in \eqref{eq:lqr-objective}.
Specifically, the optimal controller retains the same linear
feedback structure as LQR, but replaces the true state $s_t$
with its posterior estimate $\hat{s}_t$:
\begin{equation}
    a_t = -K_t \hat{s}_t,
    \qquad
    \hat{s}_t = \mathbb{E}[s_t \mid h_t],
\end{equation}
where $K_t$ is the LQR feedback gain obtained from the Riccati
recursion as in \eqref{eq:lqr-objective}. This is known as the \emph{certainty-equivalence
principle}: the controller acts as if the estimated state were
the true state~\cite{anderson2007optimal}.
Beyond the linear-quadratic-Gaussia setting, the full belief state
$p(s_t \mid h_t)$ can provide  representation of the observation
history for optimal decision-making under partial observability
\cite{kaelbling1998planning}.

\subsubsection{Unknown Dynamics: System Identification and Reinforcement Learning.}
When the transition dynamics are unknown, the agent must learn about the
environment from observed data or interactions. A classical approach is
\emph{system identification}, which aims to estimate a model of the
underlying dynamics from observed trajectories
\cite{ljung1998system}. For example, consider the above discussed fully observed linear
dynamical system $
    s_{t+1} = A s_t + B a_t + w_t,$
where $A$ and $B$ are unknown dynamics matrices and $w_t$ represents
process noise. Given trajectories of states and actions, these matrices
can be estimated through least squares regression by minimizing the
one-step state prediction error. When the states are only partially
observed, classical methods such as maximum likelihood estimation
and subspace identification can recover a state-space model from
input-output trajectories. For nonlinear systems, the dynamics can
be approximated using nonlinear parametric models, including neural
networks, learned from observed transitions.

More generally, system identification seeks to learn predictive
relationships between the agent's actions and the evolution of the
environment. Depending on the available observations and model
assumptions, this may involve predicting future states from
state-action pairs or predicting future observations from interaction
histories. The latter can be implemented using latent-state
representations that summarize past interactions. Once learned,
such models can support prediction, simulation, planning, and
model predictive control. This establishes a fundamental connection
between learning environment dynamics and model-based
decision-making.

\emph{Reinforcement learning} (RL) addresses the problem of learning
effective decision-making policies through interaction with the
environment~\cite{sutton1998reinforcement}. Its objective is to
maximize the expected cumulative reward, as formulated in
\eqref{eq:max-reward-obj}. While this formulation accommodates
multiple environments and tasks, classical RL often considers
a fixed environment and reward function, and seeks to learn
an optimal policy from experience when the environment dynamics
are unknown. Unlike system identification, which focuses on
learning a predictive model of the environment, RL directly
optimizes decision-making performance. Nevertheless, the two
are closely connected: learning the environment dynamics can
support policy optimization, although an explicit dynamics
model is not always necessary.

This distinction leads to two major approaches to RL.
\emph{Model-based RL} explicitly learns or uses a predictive
model of the environment and leverages it for planning or
policy improvement. When the
dynamics are unknown, learning such a model is closely related
to system identification, although the model may be optimized
for decision-making rather than prediction accuracy alone.
The learned model can predict future states, observations,
or rewards, allowing the agent to evaluate candidate actions
through simulated rollouts. In contrast, \emph{model-free RL}
learns policies or value functions directly from experience,
without explicitly constructing a predictive dynamics model.
Thus, both approaches aim to improve decision-making, but
differ in whether and how an environment model is explicitly
learned and used.

Although RL provides a general mathematical framework for
learning through interaction, applying it to generalist agents
introduces significant practical challenges. 
\begin{itemize}[leftmargin=*]
\item  First, many
online RL algorithms require extensive trial-and-error
interactions to acquire effective behaviors. In controlled
settings, these interactions can be generated efficiently
through simulators that support repeated trials, environment
resets, and parallel exploration. However, constructing
sufficiently accurate simulators can be difficult for complex
physical environments, particularly those involving diverse
objects, contact dynamics, and human interactions
~\cite{dulac2021challenges}. While simulation is highly
effective for certain tasks, transferring learned behaviors
to the real world introduces additional challenges due to
model mismatch and distribution shifts, known as the \emph{sim-to-real gap}
~\cite{zhao2020sim}.

\item Second, learning directly in real-world environments can be
expensive, slow, and potentially unsafe. Physical interactions
consume time and resources, while exploratory actions may
damage objects or violate safety
constraints. Moreover, many real-world situations cannot be
reset to identical initial conditions, making repeated
experimentation difficult. These limitations are particularly
important for generalist agents, which must operate across
diverse tasks and environments rather than repeatedly train
for a single task.

\item Third, conventional RL formulations often assume that the
agent can obtain reward feedback and collect trajectories
through interaction with the target environment. In contrast,
generalist agents may need to acquire transferable knowledge
from heterogeneous sources, including passive observations,
human demonstrations, previously collected interaction data,
and experience from other environments. Such data may not
contain explicit rewards or may not be generated by the
agent's current policy. Offline RL, imitation learning,
multi-task RL, and model-based approaches provide important
mechanisms for addressing aspects of these challenges
~\cite{levine2020offline,taylor2009transfer},
but efficiently integrating diverse data sources and
generalizing to unseen tasks and environments remain
central research problems.
\end{itemize}
These considerations motivate a broader perspective on
generalist-agent learning. Beyond optimizing a policy through
trial and error, an agent must determine what knowledge
to acquire, how to learn it from different forms of data,
and how to reuse it for prediction, planning, and adaptation.
This perspective connects classical system identification
and reinforcement learning to modern approaches for
learning world models, action policies, and transferable
representations.

\section{Learning to Act: Policies and World Models}
\label{sec:models-methods}

The preceding section connects generalist agents
to classical frameworks for control, state estimation,
system identification, and reinforcement learning.
These frameworks characterize different aspects of
sequential interaction, including action selection,
hidden-state estimation, and learning unknown dynamics.
Building on these foundations, we now examine how
generalist agents can learn policies and world models from data
to operate across diverse environments and tasks.

\begin{figure*}[p]
\centering
\renewcommand{\bfdefault}{bx}
\begingroup
\definecolor{ModelNavy}{HTML}{19344D}
\definecolor{ModelBlue}{HTML}{3E73A6}
\definecolor{ModelTeal}{HTML}{287E7C}
\definecolor{ModelOrange}{HTML}{AE6822}
\definecolor{ModelPurple}{HTML}{77508F}
\resizebox{0.98\textwidth}{!}{%
\begin{tikzpicture}[
  >={Latex[length=2mm,width=1.4mm]},
  every node/.style={font=\footnotesize\sffamily,text=ModelNavy,align=center},
  card/.style={rounded corners=3pt,line width=.75pt,
    minimum height=1.05cm,inner sep=5pt},
  state/.style={card,draw=ModelNavy!55,fill=white,text width=2.8cm},
  history/.style={card,draw=ModelBlue!70,fill=ModelBlue!4,text width=5.4cm},
  latent/.style={card,draw=ModelTeal,fill=ModelTeal!5,text width=3.7cm},
  observation/.style={card,draw=ModelBlue!65,fill=ModelBlue!4,
    text width=5.15cm,minimum height=1.25cm},
  latentmodel/.style={card,draw=ModelTeal!80,fill=ModelTeal!4,
    text width=5.15cm,minimum height=1.25cm},
  flow/.style={->,line width=.85pt,draw=ModelNavy!65,rounded corners=3pt},
  learned/.style={->,line width=.95pt,draw=ModelTeal},
  heading/.style={font=\small\bfseries\sffamily,anchor=west,align=left}
]

\node[heading] at (0,0)
  {(a) State, interaction history, and latent representation};
\node[state] (state) at (3.7,-1.25)
  {\textbf{State}\\[3pt]$s_t$};
\node[state] (nextstate) at (12.5,-1.25)
  {\textbf{State}\\[3pt]$s_{t+1}$};
\draw[flow] (state.east) -- node[above,font=\footnotesize\sffamily]
  {True environment dynamics\\[2pt]$p(s_{t+1}\mid s_t,a_t,e)$} (nextstate.west);

\node[history] (history) at (3.7,-3.15)
  {\textbf{Interaction history}\\[3pt]$h_t=(o_0,a_0,\ldots,a_{t-1},o_t)$};
\node[history] (nexthistory) at (12.5,-3.15)
  {\textbf{Updated history}\\[3pt]$h_{t+1}=(h_t,a_t,o_{t+1})$};
\draw[flow] (state.south) -- node[midway,right,inner sep=4pt]
  {$o_t\sim p(\cdot\mid s_t,e)$} (history.north);
\draw[flow] (nextstate.south) -- node[midway,right,inner sep=4pt]
  {$o_{t+1}\sim p(\cdot\mid s_{t+1},e)$} (nexthistory.north);
\draw[flow] (history.east) -- node[above,text width=2.5cm]
  {Append action and observation} (nexthistory.west);

\node[latent] (latent) at (3.7,-5.4)
  {\textbf{Latent representation}\\[3pt]$z_t=\phi(h_t)$};
\node[latent] (nextlatent) at (12.5,-5.4)
  {\textbf{Next latent representation}\\[3pt]$z_{t+1}=\phi(h_{t+1})$};
\draw[learned] (history.south) -- node[midway,right,inner sep=5pt]
  {Latent encoder $\phi$} (latent.north);
\draw[learned] (nexthistory.south) -- node[midway,right,inner sep=5pt]
  {Latent encoder $\phi$} (nextlatent.north);
\draw[learned] (latent.east) -- node[above,text width=4.4cm]
  {Learned latent dynamics\\[3pt]$p_\theta(z_{t+1}\mid z_t,a_t)$} (nextlatent.west);
\node[heading] at (0,-7.25)
  {(b) Model families: what is predicted, and what is given?};
\node[card,draw=ModelOrange!85,fill=ModelOrange!5,text width=7.1cm,
  minimum height=1.2cm] at (3.9,-8.4)
  {\textbf{Policy / action prediction}\\[5pt]
   $\pi_\theta(a_t\mid h_t,g)\equiv p_\theta(a_t\mid h_t,g)$};
\node[card,draw=ModelPurple!85,fill=ModelPurple!5,text width=7.1cm,
  minimum height=1.2cm] at (12.3,-8.4)
  {\textbf{Joint world--action model}\\[5pt]
   $p_\theta(a_t,o_{t+1}\mid h_t,g)$};

\node[font=\footnotesize\bfseries\sffamily,text=ModelBlue]
  at (7.0,-9.65) {Observation space};
\node[font=\footnotesize\bfseries\sffamily,text=ModelTeal]
  at (12.95,-9.65) {Latent space};

\node[text width=3.8cm] at (2.0,-10.65)
  {\textbf{Action-conditioned dynamics}\\[3pt]What if I act?};
\node[observation] at (7.0,-10.65)
  {$p_\theta(o_{t+1}\mid h_t,a_t)$};
\node[latentmodel] at (12.95,-10.65)
  {$p_\theta(z_{t+1}\mid z_t,a_t)$};

\node[text width=3.8cm] at (2.0,-12.2)
  {\textbf{Action-unconditioned dynamics}\\[3pt]What could happen?};
\node[observation] at (7.0,-12.2)
  {$p_\theta(o_{t+1}\mid h_t)$};
\node[latentmodel] at (12.95,-12.2)
  {$p_\theta(z_{t+1}\mid z_{\leq t})$};

\node[text width=3.8cm] at (2.0,-13.75)
  {\textbf{Inverse dynamics}\\[3pt]Which action led here?};
\node[observation] at (7.0,-13.75)
  {$p_\theta(a_t\mid o_{\leq t},o_{t+1})$};
\node[latentmodel] at (12.95,-13.75)
  {$p_\theta(a_t\mid z_t,z_{t+1})$};
\end{tikzpicture}%
}
\endgroup
\caption{\textbf{Policies, world models, and learned latent representations.}
(a) The underlying state $s_t$ follows the true
environment dynamics and generates observations.
A latent encoder maps the interaction history
to a learned representation $z_t=\phi(h_t)$,
whose evolution is predicted by a latent
dynamics model. Alternatively, the encoder
may use only the current observation,
$z_t=\phi(o_t)$, or observation history,
$z_t=\phi(o_{\leq t})$.
Neither direct access to $s_t$ nor a
one-to-one correspondence between $s_t$
and $z_t$ is assumed.
(b) The model families discussed in
Section~\ref{sec:models-methods}
differ in their prediction targets and
conditioning information.
Action-unconditioned dynamics models
do not condition on the current action $a_t$.
For passive data without action labels,
observation prediction takes the form
$p_\theta(o_{t+1}\mid o_{\leq t})$,
and latent models use observation-only
encodings.
For inverse dynamics prediction, latent
representations should not directly encode
the target action $a_t$ and may instead
be constructed from current observation or observation histories.
Throughout, $p_\theta$ denotes a generic
learned predictive model: the different
conditional distributions may be modeled
separately with distinct parameters or
jointly within a unified model with shared
parameters. Task conditioning may also
be included when informative.}
\label{fig:model-families}
\end{figure*}

We organize the discussion around three complementary
approaches, as illustrated in \Cref{fig:model-families},
which differ in their prediction targets, conditioning
information, and representations.
The first directly learns action policies from
demonstrations, without explicitly modeling
environment dynamics.
The second learns predictive world models in
observation or latent spaces to support planning,
control, and policy improvement.
The third jointly learns world models and action
policies, allowing world prediction and action
learning to reinforce each other.
For each approach, we discuss the underlying
model formulations, available training data and
learning objectives, and how the learned knowledge
supports decision-making.

\paragraph{Notation and conditioning conventions.}
Before proceeding, we clarify the notation
used\linebreak throughout this section. Recall that $e$ denotes the environment and $g$ denotes the task, both fixed within an
episode. The underlying dynamics depend on $e$, which is generally not directly observed by the agent. In the following
sections, we therefore omit explicit conditioning on $e$ from learned predictive models, which instead use observations or interaction histories to infer relevant environment information. We retain task conditioning in policies,
since the appropriate action generally
depends on the task. For simplicity,
individual dynamics models are typically
written without conditioning on $g$,
although task information may be included
when it provides useful predictive context.
In contrast, we retain $g$ throughout exact
joint-distribution factorizations in \Cref{sec:world-action} to ensure
consistent conditioning.

We use $p$ to denote probability distributions
induced by the underlying dynamic
 process, and $p_\theta$ to
denote learned probabilistic models
parameterized by $\theta$.
Similarly, we use $\pi$ to denote a generic
action policy and $\pi_\theta$ to denote
a parameterized policy. For simplicity, we use $\theta$ as a generic
parameter symbol and distinguish different
models by their predicted variables and
conditioning information.
These models need not share the same parameters, and
they may be implemented by separate neural
networks or jointly represented within a
unified architecture.
Our discussion focuses primarily on the
conditional distributions being modeled,
rather than their specific parameterizations
or learning algorithms.

Suppose the agent has access to interaction
trajectories collected from one or more behavior
policies, such as human demonstrations or 
teleoperated robotic actions. Each trajectory consists of
a task specification $g$ and a sequence of
observations and actions
$(o_0,a_0,o_1,\ldots,a_{T-1},o_T)$.
For simplicity, we use a common horizon $T$,
although trajectories may have different
lengths and need not contain complete episodes.
The demonstrated behaviors need not be optimal,
but are assumed to provide useful examples
of how to accomplish the tasks of interest. The same interaction trajectories can provide
supervision for different prediction tasks.
In particular, we consider the following
datasets for action prediction and
action-conditioned dynamics prediction:
\begin{equation}
\begin{aligned}
\Dact
&=
\left\{
(h_t,g,a_t)
\right\}, \quad
\Ddyn =
\left\{
(h_t,a_t,o_{t+1})
\right\},
\end{aligned}
\label{eq:interaction-datasets}
\end{equation}
where $h_t=(o_0,a_0,\ldots,a_{t-1},o_t)$
is the interaction history preceding
action $a_t$, and the examples are extracted
over $0\leq t<T$ from the available trajectories. We omit $g$ from $\Ddyn$ under the assumption
that the underlying environment dynamics do not
directly depend on the task given the action. However, when the
task specification provides additional information
about the environment, it may also be included
as a conditioning variable in dynamics prediction. 

In addition, passive observational data
can be used for learning predictive models
without requiring action labels.
We denote the corresponding dataset by
\begin{equation}
\Dpass
=
\left\{
(o_{\leq t},o_{t+1})
\right\},
\label{eq:passive-dataset}
\end{equation}
where $o_{\leq t}=(o_0,\ldots,o_t)$
denotes the observation history.
Unlike $\mathcal D_{\mathrm{act}}$ and
$\mathcal D_{\mathrm{dyn}}$, this dataset
does not require the actions associated
with the observed transitions and can
therefore be constructed from passive
observation sequences, such as videos,
as well as from interaction trajectories
with action labels omitted.

\subsection{Direct Policy Learning from Demonstrations}
We begin with the most direct approach to learning
an agent: learning an action policy from demonstrations
without explicitly modeling the environment dynamics.
Recall from \Cref{sec:generalist-agents} that a policy
$\pi(a_t\mid h_t,g)$ specifies a conditional
distribution over actions given the interaction
history and task.
Given the action examples in $\Dact$, defined
in \eqref{eq:interaction-datasets}, we seek to
learn a parameterized policy
$\pi_\theta(a_t\mid h_t,g)$, where $\theta$
denotes the model parameters.

A standard approach is \emph{behavior cloning},
which formulates policy learning as a supervised
learning problem by maximizing the likelihood
of demonstrated actions~\cite{osa2018algorithmic}:
\begin{equation}
\max_{\theta}
\;\sum_{(h_t,g,a_t)\in\Dact}
\log \pi_\theta(a_t\mid h_t,g).
\label{eq:policy-MLE}
\end{equation}
From a predictive modeling perspective, the
learned policy can be viewed as an
\emph{action model} that predicts actions
conditioned on the interaction history and task. This connects policy learning to other
conditional prediction problems, including
the world-model learning approaches discussed later. In practice, such policies can be
implemented using different architectures,
including Transformers and diffusion models,
with corresponding learning objectives. When demonstrations are generated by a single behavior policy
$\pi$, the learned policy $\pi_\theta$ aims to approximate
$\pi$ on histories represented in the training data. More generally, 
modern datasets often combine demonstrations from
multiple human operators, robots, or algorithms. In this case, the learned policy approximates the conditional action
distribution induced by the combined dataset, which may reflect a mixture of different behavior
policies.

Many recent agent models can be understood as extensions
of this direct supervised policy learning paradigm.
They differ primarily in the information used to condition
the policy, the parameterization of the action distribution,
and the underlying model architecture, while sharing the
common goal of learning actions directly from demonstrations. A common approach is to parameterize the policy autoregressively by decomposing the conditional action distribution into a sequence of action tokens or action chunks, analogous to next-token prediction in language models. Decision Transformer~\cite{chen2021decision} formulates offline reinforcement learning as conditional sequence modeling by predicting actions conditioned on past observations, actions, and a desired return-to-go. More recent robotic foundation models, including RT-1~\citep{brohan2023rt1}, RT-2~\citep{brohan2023rt2}, and OpenVLA~\citep{kim2025openvla}, similarly represent robot actions as sequences of discrete tokens and train the policy using autoregressive maximum likelihood. Despite their architectural differences and conditioning information, these methods all optimize the policy directly through supervised learning rather than explicitly modeling the environment dynamics, learning a value function, or directly maximizing cumulative reward. For continuous and multimodal action spaces, the policy can instead be parameterized by a conditional diffusion model. Diffusion Policy~\citep{chi2025diffusion} represents the policy as a conditional denoising diffusion process that generates action sequences conditioned on the interaction history. Although the training objective is a denoising (or score-matching) loss instead of the action-level log-likelihood in standard behavior cloning, it remains a form of supervised policy learning. The diffusion parameterization is particularly effective for continuous, high-dimensional, and multimodal action distributions.

Despite their effectiveness, direct supervised policy learning
faces several important limitations. First, learning a broadly
capable policy requires demonstrations covering a sufficiently
diverse range of environments, tasks, embodiments, and behaviors.
Second, the learned policy is trained on histories generated by
the behavior policies but is deployed on histories induced by
its own actions. Prediction errors can therefore cause the agent
to encounter situations poorly represented in the training data,
potentially leading to compounding errors over long horizons~\cite{ross2011reduction}. Third, because behavior cloning
optimizes agreement with demonstrated actions rather than
cumulative reward, its performance depends strongly on the
quality and coverage of the demonstrations, with no explicit
mechanism for improving beyond the demonstrated behavior.

To partially address these limitations, many recent
VLA models leverage knowledge acquired
through large-scale multimodal pretraining. For example,
RT-2 and OpenVLA
build on pretrained vision-language models and adapt them to
robotic control using action-labeled demonstrations.
The pretrained models provide rich semantic and visual
representations learned from large-scale image-text data,
while robotic demonstrations ground these representations
in executable actions. By transferring knowledge from
multimodal pretraining, these approaches can improve
generalization to new instructions, objects, and visual
configurations, reducing reliance on task-specific
robot demonstrations.

Nevertheless, pretrained representations alone do not
eliminate the limitations of direct policy learning.
In particular, these policies do not explicitly model
how the environment evolves in response to actions,
making it difficult to predict the consequences of
alternative behaviors or systematically improve through
planning. This motivates approaches that learn
\emph{predictive world models}, which can support
reasoning about future states, evaluating candidate
actions, and improving policies beyond direct imitation.

\subsection{Learning World Models for Planning and Control}

Direct policy learning predicts actions from the
interaction history and task, without explicitly
modeling how the environment responds to those
actions. An alternative is to learn a
\emph{world model} that captures knowledge
about the environment to support prediction,
planning, or policy improvement. As discussed in the introduction, the notion
of a world model is broad and need not be
restricted to predicting environment dynamics.
Such models may capture different aspects
of the world, including its structure,
states, relationships, and dynamics, at
different levels of abstraction.
In this section, we focus on a particularly
important class of world models,
\emph{dynamics models}, which predict how
observations or learned representations
evolve over time, either with or without
conditioning on actions.
We introduce observation-space and latent-space
formulations and discuss how they can support
decision-making.

\subsubsection{Observation-Space World Models}

As discussed in the preceding section, under partial
observability, the underlying environment state is unavailable, and system identification can instead be formulated as predicting future observations from the interaction history and actions. This motivates action-conditioned observational world models, which learn the observable consequences of actions without
explicitly modeling the underlying state transitions~\cite{finn2016unsupervised}.

\begin{definition}[Action-conditioned observational dynamics model]
{An \emph{action-conditioned observational dynamics model} is a probabilistic
model $p_\theta(o_{t+1}\mid h_t,a_t)$ that predicts the next
observation conditioned on the interaction history $h_t$ and
current action $a_t$.}
\end{definition}
As in behavior cloning, $\theta$ denotes
the parameters of the model being learned,
without implying that the dynamics model
and policy share parameters. When the task specification provides
additional information about the environment,
the model may also condition on $g$.
For simplicity, we omit this conditioning
in the following discussion. The goal is to approximate the true conditional
distribution of future observations.
For a generalist agent operating across
different environments, this distribution
can be expressed as
\begin{equation}
\begin{aligned}
p_\theta(o_{t+1}\mid h_t,a_t)
&\approx p(o_{t+1}\mid h_t,a_t)\\
&=
\int
p(o_{t+1}\mid s_{t+1},e)\,
p(s_{t+1}\mid s_t,a_t,e)
p(s_t,e\mid h_t,a_t)\,
ds_t\,ds_{t+1}\,de.
\end{aligned}
\label{eq:observational-dynamics}
\end{equation}
Here, $p(s_t,e\mid h_t,a_t)$ captures uncertainty
about both the underlying state and the
environment, which the agent must infer from its interaction history.  An action-conditioned
world model can predict the consequences
of actions, supporting
planning, model predictive control, and
model-based reinforcement learning.

The observational dynamics model can be
learned from the action-conditioned transition
examples in $\Ddyn$, defined
in \eqref{eq:interaction-datasets}, where
$(h_t,a_t)$ serves as the input and
$o_{t+1}$ as the prediction target.
A standard approach is to maximize the
likelihood of observed transitions:
\begin{equation}
\max_{\theta}\;
\sum_{(h_t,a_t,o_{t+1})\in\Ddyn}
\log p_\theta(o_{t+1}\mid h_t,a_t).
\label{eq:observation-prediction}
\end{equation}
Different model architectures and learning
objectives can be used for observation
prediction. Deterministic models predict
a single observation $\widehat o_{t+1}$
and typically minimize a regression loss
against the observed outcome
~\cite{finn2016unsupervised}. More expressive approaches can be used to 
capture stochasticity and multimodal
future outcomes~\cite{babaeizadeh2017stochastic,rigter2024avid}.

However, collecting large-scale interaction data with
action annotations can be expensive, particularly in
physical environments. In contrast, passive observations,
such as Internet-scale videos, are much more abundant.
This motivates learning world models from observations alone, without requiring the actions that generated them.

\begin{definition}[Action-unconditioned observational dynamics model]
An \emph{action-unconditioned observational
dynamics model} predicts the next observation
without conditioning on the current action
$a_t$. In a probabilistic formulation, it
can be represented as
$p_\theta(o_{t+1}\mid h_t)$,
where the interaction history $h_t$ may
include past observations and actions.
When only observations are available,
the model instead takes the form
$p_\theta(o_{t+1}\mid o_{\leq t})$,
where $o_{\leq t}=(o_0,\ldots,o_t)$.
\label{def:action-uncond-obs-dynamic}
\end{definition}
The key distinction from action-conditioned
dynamics models is whether the current action
$a_t$ is included in the conditioning context,
rather than whether past actions are available.
As before, we use $p_\theta$ for different
predictive models, with the conditioning
variables distinguishing their formulations,
without implying that they share parameters.

In the visual domain, the observation-only
formulation is commonly studied as
\emph{video prediction}, where models predict
future video frames from preceding frames~\cite{gao2022simvp}.
Such \emph{vidoe predition models} can be learned from the passive
observation dataset $\Dpass$, defined in
\eqref{eq:passive-dataset}, without requiring
action or task labels.
Each pair $(o_{\leq t},o_{t+1})$ provides
a prediction context and its target.
A standard approach is to maximize the
likelihood of observed transitions:
\begin{equation}
\max_{\theta}\;
\sum_{(o_{\leq t},o_{t+1})\in\Dpass}
\log p_\theta(o_{t+1}\mid o_{\leq t}).
\label{eq:passive-observation-prediction}
\end{equation}
Other generative learning objectives can
also be used. Representative approaches
include autoregressive video prediction,
stochastic video prediction
~\cite{babaeizadeh2017stochastic},
and diffusion-based video generation
~\cite{ho2022video}.
These approaches employ different
architectures and training objectives,
including maximum likelihood, variational
learning, and denoising-based objectives.
By predicting future observations, they
can learn temporal regularities, object
interactions, and aspects of physical
and semantic structure from large-scale
observational data.

Unlike action-conditioned dynamics models,
action-unconditioned predictors implicitly
average over the actions or behaviors
represented in the data, limiting their use
for planning and control.
Nevertheless, they can provide valuable
prior knowledge about how the world evolves. Such knowledge
may be transferred to action-conditioned  models
or policies through subsequent training on  interactive trajectories, potentially reducing the
amount of action-labeled data required.

\subsubsection{Latent-Space World Models}

For learning a generalist agent that maximizes the cumulative
reward in \eqref{eq:max-reward-obj}, accurately predicting future
observations may be unnecessary. Instead, an alternative approach
is to learn a latent representation that preserves information
relevant to future prediction and decision-making, together with
a predictive model that describes how this representation evolves
over time~\cite{hafner2019learning,schrittwieser2020mastering,
hansen2022temporal}.

\begin{definition}[Latent encoder]
A \emph{latent encoder} $\phi$ is a mapping that maps the interaction history
\(
h_t = (o_0, a_0, \ldots, a_{t-1}, o_t)
\)
to a latent representation \(
z_t=\phi(h_t).
\)
The encoder summarizes
information relevant to future prediction and decision-making.
\end{definition}

In practice, the encoder may operate on the current
observation alone, $z_t=\phi(o_t)$, as in
JEPA-style representation-learning methods
~\cite{lecun2022path,assran2023self,bardes2024revisiting},
or incorporate the interaction history to account
for partial observability. For example, a recurrent
encoder can update the latent representation as $
z_t=\phi(z_{t-1},a_{t-1},o_t),$
as in recurrent state-space world models
~\cite{hafner2019learning,hafner2019dream}.
The latter can be viewed as a recursive implementation
of a history-dependent encoder.

\begin{remark}[Relation to the underlying state $s$]
The latent representation $z_t$ should not be confused
with the underlying state $s_t$. As illustrated in \Cref{fig:model-families}(a),
the state $s_t$ is part of the environment and evolves
according to the transition dynamics
$p(s_{t+1}\mid s_t,a_t,e)$, whereas $z_t$ is a learned
representation constructed from the agent's observations
and interaction history. In fully observable environments, $z_t$ may approximate
the underlying state or a lower-dimensional abstraction
of it. Under partial observability, however, the current
observation generally does not contain sufficient information
to determine the underlying state. A history-dependent
latent representation may therefore approximate the belief
state $p(s_t\mid h_t,g)$ or encode other predictive information
relevant to future outcomes and decisions.
The pair $(z_t,g)$ must retain the relevant
information if it is to replace $(h_t,g)$. Consequently,
there need not be a one-to-one correspondence between
$z_t$ and $s_t$.
\end{remark}

A latent representation alone does not specify how the
environment evolves. To model its evolution, we introduce
latent dynamics models that predict future latent
representations, either with or without conditioning on
the agent's actions. 

\begin{definition}[Latent dynamics models]
An \emph{action-conditioned latent dynamics model}
is a probabilistic model
$p_\theta(z_{t+1}\mid z_t,a_t)$
that predicts the next latent representation
conditioned on the current latent representation
and action.
An \emph{action-unconditioned latent dynamics model}
instead predicts future latent representations
without conditioning on actions, represented by
$p_\theta(z_{t+1}\mid z_{\leq t})$.
\end{definition}
As with observational dynamics models,
we use the same notation $p_\theta$
for both formulations, without implying
that they share parameters.
For passive dynamics learning, latent
representations are constructed from
observations alone, such as
$z_t=\phi(o_t)$ or
$z_t=\phi(o_{\leq t})$,
since action labels are unavailable.
For simplicity, we omit task conditioning,
although it can be included when useful.

Given a fixed encoder $\phi$, learning latent dynamics is similar to
learning observational dynamics, except that prediction takes place in a latent representation
space. Specifically,
interaction examples in $\Ddyn$ can be
mapped to latent transitions
$(z_t,a_t,z_{t+1})$, providing supervision
for action-conditioned latent dynamics.
Similarly, passive observations in $\Dpass$
can support latent prediction without action
labels. As for learning observational dynamics, probabilistic models may be trained
through maximum likelihood or other generative
objectives, while deterministic models typically
use regression or representation-prediction
losses~\cite{hafner2019learning,
hansen2022temporal,assran2023self}.
Representative passive approaches include
V-JEPA~\cite{bardes2024revisiting,assran2025v},
which learns predictive representations from
videos without action annotations or
pixel-level reconstruction.

In practice, the latent encoder and dynamics
model are often learned jointly rather than
sequentially. This introduces an important
challenge: accurate latent prediction alone
does not guarantee that the learned
representation captures meaningful or
decision-relevant information about the
environment. For example, a constant encoder
$z_t=\phi(h_t)=c$ admits perfect prediction
of $z_{t+1}=c$, but provides no useful
information for decision-making.
More generally, the learned representation
may preserve easily predictable features
while discarding information important
for planning and control.
This motivates the notion of
\emph{control sufficiency}, discussed next.

\begin{definition}[Control-sufficient latent representation]
A latent representation $z_t=\phi(h_t)$ is
\emph{control sufficient} if it preserves the information
necessary for optimal decision-making. Specifically,
an agent operating on the latent representation can
achieve the same optimal performance as an agent
operating on the full interaction history in the
original decision-making problem \eqref{eq:max-reward-obj}.
\end{definition}
This definition is
closely related to several classical theories of state
representation. In fully observable MDPs,
\emph{state abstraction} seeks lower-dimensional
representations that preserve value functions or optimal
policies~\cite{li2006unified}. MDP
\emph{homomorphisms} characterize abstractions that
preserve transition and reward structures, thereby
inducing equivalent planning problems in abstract state
spaces~\cite{ravindran2003smdp}. Closely related,
\emph{bisimulation} characterizes states that are
behaviorally indistinguishable by requiring equivalent
rewards and transition behavior under each action
~\cite{givan2003equivalence,ferns2004metrics}.
For partially observable environments,
predictive state representations
represent the state through predictions of future
observable events rather than unobserved physical
states~\cite{littman2002predictive}.

Modern latent-space world models can be understood
as learning approximate predictive representations
using neural networks. Reconstruction-based models,
such as PlaNet and Dreamer, learn latent dynamics
together with observation reconstruction objectives
~\cite{hafner2019learning,hafner2019dream}.
JEPA-style approaches predict future latent
representations without requiring pixel-level
reconstruction~\cite{lecun2022path,assran2023self,bardes2024revisiting}.
A key challenge in such approaches is representation
collapse, where the encoder maps different observations
to nearly identical representations.
Recent work, such as LeWorldModel,
addresses this issue by regularizing the latent
representations toward an isotropic Gaussian
distribution, enabling stable end-to-end learning
of action-conditioned latent dynamics
~\cite{maes2026leworldmodel}.
Control-oriented approaches instead
learn latent dynamics together with decision-relevant
quantities such as rewards or values, without requiring
full observation reconstruction
~\cite{hansen2022temporal}.
The resulting latent dynamics models can support
inference-time planning, including model predictive
control, or provide predictive rollouts for policy
optimization through reinforcement learning.

\subsubsection{Learning Policies from Predictive Models}

Once a predictive world model has been learned, either in observation space or in latent space, it can be used to construct a policy for the original decision-making objective in \eqref{eq:max-reward-obj}. A common approach follows the principle of certainty equivalence, treating the learned model as an approximation to the true environment and optimizing decisions accordingly. More generally, model-based methods may also account for prediction uncertainty and model errors. Depending on how the predictive model is used, we distinguish three approaches: inference-time planning, policy learning through simulated experience, and control from passive predictive models. For simplicity, we illustrate the main ideas using latent dynamics, although analogous principles apply to observation-space models.

\paragraph{Inference-time Planning}
A first approach is to select actions through
online planning with the learned dynamics model.
At each time step, the agent predicts the outcomes
of candidate action sequences and selects actions
that optimize a desired objective.

For example, consider a goal-reaching task,
where $z_g$ denotes the latent representation
of a desired goal, obtained using a compatible
encoder or goal representation.
Starting from the current latent representation
$z_t=\phi(h_t)$, the agent may plan over
a horizon of $H$ steps by solving
\begin{equation}
\begin{aligned}
\min_{a_t,\ldots,a_{t+H-1}}
&\quad
\mathbb E\left[
    \dist(z_{t+H},z_g)
\right]\\
\text{s.t.}\quad
&z_{k+1}\sim p_\theta(\cdot\mid z_k,a_k),
\quad k=t,\ldots,t+H-1.
\end{aligned}
\label{eq:latent-mpc}
\end{equation}
Here, $\dist$ measures the distance to the
goal in latent space, and the expectation
is taken over trajectories generated by
the learned dynamics model $p_\theta$.

Rather than executing the entire planned action
sequence, the agent typically executes only the
first action and replans after receiving a new
observation. This receding-horizon strategy, known
as MPC, continually
incorporates feedback to mitigate prediction errors
and environmental disturbances~\cite{camacho2007constrained}.
Learned predictive models have enabled MPC in high-dimensional visual environments~\cite{ebert2018visual} and latent-space planning~\cite{hafner2019learning,maes2026leworldmodel}.
Other planning approaches, such as MuZero, combine
learned latent dynamics with tree search to select
actions~\cite{schrittwieser2020mastering}.

\paragraph{Policy Learning through Simulated Experience.}

Instead of planning from scratch at every decision
step, another approach is to use the learned
predictive model as a simulator for policy learning.
By generating synthetic trajectories through the
learned dynamics, the agent can optimize a
parameterized policy using reinforcement learning algorithms~\cite{sutton1998reinforcement,chi2025statistical}
without requiring an equivalent number of
interactions with the real environment.

Importantly, policy optimization need not depend
on the internal structure of the learned world model.
Standard reinforcement learning methods, such as
policy gradient, actor-critic, or value-based learning,
can be applied to simulated trajectories
~\cite{sutton1991dyna,janner2019trust}.
For example, given a learned latent dynamics model
$p_\theta(z_{t+1}\mid z_t,a_t)$ and a policy
$\pi_\theta(a_t\mid z_t,g)$, the agent can generate
simulated trajectories according to
\begin{equation}
\begin{aligned}
a_t &\sim \pi_\theta(\cdot\mid z_t,g),\quad
z_{t+1} &\sim p_\theta(\cdot\mid z_t,a_t).
\end{aligned}
\label{eq:model-rollout}
\end{equation}
To optimize the policy, simulated trajectories
must also provide a reward signal, which can
be obtained from a learned reward predictor
or an appropriate task-specific evaluator such as using VLM.
Reinforcement learning can then improve the
policy using simulated experience, rather than
merely imitating demonstrated actions.

Recent work extends this paradigm to reinforcement
fine-tuning of pretrained VLA policies~\cite{li2025vla,zhu2026wmpo,jiang2026wovr}.
These approaches demonstrate the potential of
world models to reduce costly real-world interactions
during policy improvement, while highlighting
the importance of accurate simulated dynamics
for reliable policy optimization. Alternatively, when the learned dynamics are
differentiable, policy optimization can exploit
gradients propagated through predicted state
transitions, allowing the model to participate
directly in the optimization process~\cite{hafner2019dream}. In both cases, optimization is amortized into
the policy parameters, allowing actions to be
selected directly at inference time without
repeatedly solving a planning problem.
These approaches can potentially improve
beyond the behavior represented in demonstration
data by evaluating alternative actions in the
learned model. However, their effectiveness
depends on the accuracy of the predictive model,
particularly for trajectories outside the
training distribution.

\paragraph{Control from Action-unconditioned Predictive Models.}
The preceding approaches rely on action-conditioned
predictive models that estimate the consequences
of candidate actions. In contrast, action-unconditioned
predictive models learn how observations or
latent representations evolve without access
to the actions responsible for the transitions.
Consequently, predicted future trajectories
do not directly specify how an agent should
act to realize them.

One approach to bridging this gap is to learn
an \emph{inverse dynamics model} that infers the
action associated with a transition between
latent representations~\cite{pathak2017curiosity,baker2022video}.
This allows agents to leverage predictive
world models learned from abundant passive
observations, while using relatively limited
action-labeled data to learn how to translate
predicted transitions into executable actions.

\begin{definition}[Inverse dynamics model]
An \emph{inverse dynamics model} predicts
the action associated with a transition: in observation space,
it can be represented as
$p_\theta(a_t\mid o_{\leq t},o_{t+1})$, while in latent space, it can be represented by
a probabilistic model
$p_\theta(a_t\mid z_t,z_{t+1})$.
\end{definition}
Both formulations can be learned from
action-labeled transitions in $\Ddyn$,
using $a_t$ as the prediction target. Once learned, an inverse dynamics model can
serve several purposes. First, given a trajectory
predicted by a passive world model, it can
infer candidate actions corresponding to
successive transitions, potentially enabling
the agent to execute the predicted trajectory.
Second, it can infer missing action labels
from passive observation sequences in $\Dpass$,
thereby expanding the available action-labeled
training data for learning action-conditioned
dynamics models or policies~\cite{pathak2017curiosity,baker2022video}. However, passive predictions are not
necessarily controllable: some predicted
transitions may be infeasible, and multiple
actions may produce similar outcomes.
Consequently, inverse dynamics requires
sufficient action-grounded supervision,
and effective control may need additional
feedback or feasibility constraints.
An alternative is to infer latent action
representations from passive observations.
For example, Genie learns latent actions
from unlabeled videos and uses them to
condition a generative dynamics model,
enabling controllable video prediction
without explicit action labels~\cite{bruce2024genie}.
This illustrates how passive observations
can support learning interactive world models,
although grounding learned latent actions
in executable commands remains an
additional challenge.

\subsection{World-Action Models: Joint Learning of World Models and Action Policies}
\label{sec:world-action}

The previous subsections introduce two
complementary approaches: learning action policies
directly from demonstrations, and learning world
models, particularly dynamics models, to support
planning and policy improvement.
Although these approaches emphasize different
learning objectives, world modeling and policy
learning are closely related.
A world model captures knowledge about the
environment, while a policy selects actions
to achieve a task.
Even when a policy is learned without an
explicit world model, effective action selection
may require implicitly capturing aspects of
the environment and its dynamics.
This motivates a third approach:
jointly learning world models and action
policies so that predictive modeling and
action learning can reinforce one another. Such approaches are often referred to as
\emph{world-action models}
~\cite{lin2026jepa,ye2026world}.

\paragraph{Joint Learning through Shared or
Interacting Representations.}

A general perspective on jointly learning
world models and policies is to consider
the joint conditional distribution of
future observations and actions,
$p(o_{t+1},a_t\mid h_t,g)$.
This distribution captures the relationship
between actions and their observed consequences,
without prescribing how they should be
predicted. By the chain rule of probability,
the joint distribution admits two complementary
factorizations:
\begin{equation}
\begin{aligned}
p(o_{t+1},a_t\mid h_t,g)
&=
\underbrace{\pi(a_t\mid h_t,g)}_{
\text{Policy}}
\underbrace{p(o_{t+1}\mid h_t,a_t,g)}_{
\text{Action-conditioned dynamics}}
\\
&=
\underbrace{p(o_{t+1}\mid h_t,g)}_{
\text{Future prediction}}
\underbrace{p(a_t\mid h_t,o_{t+1},g)}_{
\text{Inverse dynamics}}.
\end{aligned}
\label{eq:joint-world-action-factorization}
\end{equation}
The first factorization predicts actions
and their consequences, connecting policy
learning with action-conditioned dynamics
modeling. The second predicts future
observations followed by their associated
actions, connecting future prediction with
inverse dynamics. The
future-prediction term
$p(o_{t+1}\mid h_t,g)$ is an instance
of action-unconditioned observational
dynamics modeling in
\Cref{def:action-uncond-obs-dynamic}. Here, we retain task conditioning in the general
factorization for completeness, although
it is omitted from the dynamics models
introduced earlier for simplicity.

Both factorizations describe the same
joint distribution but suggest different
ways of organizing learning and generation.
The same principles extend to latent
representations and longer sequences
of future observations and actions.
More broadly,
\eqref{eq:joint-world-action-factorization}
suggests different approaches to learning
the joint distribution: modeling it directly
or learning its conditional factors.
These approaches may use separate models
or a unified architecture with shared
parameters.
This probabilistic perspective connects
policy learning with the different dynamics
models introduced earlier, while allowing
flexibility in learning objectives,
architectural designs, and mechanisms
for integrating world prediction and
action generation.

\begin{itemize}[leftmargin=*]
\item \emph{Joint learning objectives.}
One approach explicitly combines policy
learning with a world-prediction objective.
For example, a shared representation
$z_t=\phi(h_t)$ can support
both an action predictor
$\pi_\theta(a_t\mid z_t,g)$
and an observational dynamics model
$p_\theta(o_{t+1}\mid z_t,a_t)$,
where the latter may alternatively predict
future latent representations.
The two components can be trained jointly
using action-labeled interaction trajectories,
which provide supervision for both action
prediction and future observation prediction.
The datasets $\Dact$ and $\Ddyn$ provide
the corresponding training examples,
with task labels used for policy learning.
Such joint learning encourages representations
that capture environmental dynamics while
preserving information relevant to
decision-making~\cite{wu2024unleashing,lin2026jepa}. Alternatively, a unified generative model
can learn the joint distribution of future
observations and actions without explicitly
separating policy and world-prediction losses~\cite{ye2026world}.
These approaches illustrate different
ways to integrate world modeling and
action generation through shared
representations or generative objectives.

\item {\it Learning signals}. Recall that direct policy learning estimates
$\pi_\theta(a_t\mid h_t,g)$ from demonstrations,
typically by minimizing the supervised learning
objective in \eqref{eq:policy-MLE}.
In many robotic settings, the action space
is relatively low-dimensional. For example,
a robotic manipulator may use a 7-dimensional
action representation consisting of a
6-DoF end-effector pose and a gripper command.
Consequently, each training example provides
supervision over relatively few action variables.
In contrast, visual observations, such as
images and video frames, are high-dimensional,
often containing hundreds of thousands or
millions of pixel values.
Predicting future observations therefore
provides a much higher-dimensional learning
signal per training example.
Jointly learning action and observation
prediction allows the model to leverage
both low-dimensional action supervision
and high-dimensional visual prediction signals.

\item \emph{Architectural realization.}
The learning objectives above can be implemented
using different degrees of parameter sharing
and interaction. A common design uses a shared
encoder or Transformer backbone with
modality-specific prediction heads, allowing
world prediction and action prediction to
shape common representations. Other designs
use separate video and action modules that
exchange information through attention,
as in Mixture-of-Transformers (MoT)
architectures~\cite{yuan2026fast,wang2026openwam}. The attention mechanism can support different
conditional dependencies. For example, allowing
video tokens to attend to action tokens enables
action-conditioned world prediction, whereas
allowing action tokens to attend to future-video
representations enables future-conditioned
action prediction. Bidirectional interaction
can support joint video-action generation.
The precise probabilistic interpretation
depends not only on attention patterns,
but also on the training objective and
generation procedure.

These architectural choices are not mutually
exclusive. For example, DreamZero combines
a shared generative backbone with
modality-specific output components.
Thus, the essential distinction is how
information is shared between world prediction
and action generation.
\end{itemize}

\paragraph{Training versus Inference.}
Joint learning of world models and policies
does not necessarily require explicit world
prediction at inference time.
Depending on the model architecture and
inference procedure, a jointly trained model
may directly generate actions, jointly predict
future observations and actions, or use
predicted futures for planning.
For example, Fast-WAM~\cite{yuan2026fast}
shows that a model jointly trained can generate actions
without explicitly predicting future frames
at inference time, achieving competitive
performance with approaches that jointly
generate future observations and actions.
This suggests that the benefits of world
modeling may arise from acquiring useful
representations during training, rather
than necessarily requiring explicit future
prediction during inference.
Understanding the respective contributions
of joint learning during training and
joint prediction during inference remains
an important direction for future research.

\subsection{Language Models as Implicit World Models
and Generalist Agents}

Most of the approaches discussed so far focus on
physical intelligence, particularly robotics and
embodied agents, where world models aim to capture
environmental dynamics and the consequences of
physical actions. However, the principles of world
modeling extend beyond physical environments, and
acquiring knowledge about the world does not
necessarily require an explicit world-modeling
objective. More generally, models trained to
predict sequences of observations, actions, and
other information may implicitly acquire knowledge
about environmental structure and dynamics.
LLMs provide an
interesting example of this possibility.

LLMs are primarily pretrained on large-scale
text corpora through next-token prediction.
Although this objective does not explicitly
target environment dynamics, the training
data contain extensive descriptions of
events, interactions, actions, and their
consequences. By learning to predict such
sequences, LLMs may develop implicit
representations of environmental states,
temporal relationships, and action outcomes.
More importantly, when observations and
actions are represented as sequences of
tokens, the same autoregressive framework
can learn both world prediction and action
generation through different conditional
predictions. This reveals a connection
between language modeling, world modeling,
and policy learning: these capabilities
can be acquired through a unified predictive
learning framework, without requiring
separate models or explicitly distinct
learning objectives.

\paragraph{Implicit World Modeling through
Next-Token Prediction.}

The standard objective of autoregressive
language modeling is to learn the joint
distribution of a token sequence
$(x_0,\ldots,x_n)$ through next-token
prediction. By the chain rule of probability,
this distribution can be factorized as
\begin{equation}
p(x_0,\ldots,x_n)
=
\prod_{i=0}^{n}
p(x_i\mid x_{0},\ldots,x_{i-1}).
\end{equation}
An autoregressive language model learns
these conditional distributions by
predicting each token from its preceding
context.

The same principle can be extended to
interaction trajectories.
Consider a sequence of observations and
actions serialized as
$(g,o_0,a_0,o_1,a_1,\ldots)$,
where observations and actions are
represented as sequences of tokens,
and the task $g$ is included in the context.
At the level of complete observations
and actions, autoregressive prediction
corresponds to learning two conditional
distributions, $\pi_\theta(a_t\mid h_t,g)$ and $p_\theta(o_{t+1}\mid h_t,a_t,g)$. This corresponds to the first factorization
of the joint world-action distribution
in \eqref{eq:joint-world-action-factorization}.
Both conditionals can be
learned through the same token-level
next-token prediction objective, without
requiring separate networks or learning
mechanisms. Thus, autoregressive modeling
of interaction trajectories can jointly
provide supervision for policy learning
and world modeling.

Alternatively, consider a different
serialization in which the future observation
$o_{t+1}$ precedes the corresponding action
$a_t$, given the history $h_t$ and task $g$, i.e., $(g,\ldots,o_{t+1},a_{t},\ldots)$.
Autoregressive prediction then corresponds
to learning $p_\theta(o_{t+1}\mid h_t,g)$ and $p_\theta(a_t\mid h_t,o_{t+1},g)$, which correspond to the second factorization
of \eqref{eq:joint-world-action-factorization}.
By changing the serialization order and
conditioning context, autoregressive models
can therefore learn different conditional
distributions underlying world modeling
and action prediction.

More broadly, this perspective highlights
a connection between language modeling,
world modeling, and policy learning:
all can be formulated as conditional
prediction problems, with their roles
determined by the prediction targets
and available context.
This perspective suggests that next-token
prediction over sufficiently diverse data
may implicitly learn a mixture of predictive
models, including models of actions, dynamics,
and interactions. Evidence for emergent
world representations has been observed in
language models trained on game sequences,
such as Othello-GPT~\cite{li2022emergent}.
However, predicting plausible sequences does
not necessarily imply learning accurate
action-conditioned dynamics. In particular,
standard language pretraining does not
guarantee reliable predictions under
unfamiliar interventions or over long
interaction horizons
~\cite{wang2024can}.

\paragraph{Explicit World Modeling through
Interaction-Based Fine-Tuning.}

Although next-token prediction on general-purpose
text may enable LLMs to acquire implicit world
knowledge, their ability to predict the consequences
of actions can be further improved through
explicit world-modeling supervision on interaction
trajectories. 
Recent work follows two complementary approaches.
The first treats the \emph{world model as a separate
simulator} for predicting action consequences and
supporting planning, as
demonstrated by Web Agents with World Models,
WebDreamer, WebEvolver, and WebWorld
~\cite{chae2025web,gu2025webdreamer,
fang2025webevolver,xiao2026webworld}.
Reinforcement World Model Learning further
explores reinforcement learning to improve
the accuracy of predicted transitions
~\cite{yu2026reinforcement}. The second approach \emph{integrates world
modeling into the policy itself}, grounding
action selection in learned environment dynamics
without requiring a separate simulator~\cite{lu2026policy,zhang2026earlyexperience}.
This approach is closely related to
world-action models, where world modeling
and action generation are learned within
a unified model. Both approaches
can use similar interaction data and
prediction objectives. This reinforces
the view that world modeling and policy
learning are distinct but closely related
capabilities that need not be implemented
as separate models.

More broadly, LLM-based agents suggest a promising
path toward generalist intelligence, where
large-scale pretraining provides broad world
knowledge, and interaction-based learning
progressively grounds and refines this knowledge, enabling agents to better predict, adapt to,
and act across diverse environments and tasks.

\section{Discussion and Conclusion}
\label{sec:discussion-conclusion}
In this paper, we have introduced world models
from the perspective of learning generalist agents,
connecting them to classical sequential decision-making
and different approaches to policy learning.
We have highlighted how world modeling and
policy learning can be understood through a
unified conditional prediction perspective,
without necessarily requiring separate models.
Building on these connections, we conclude
by discussing several important challenges
and future directions toward developing
generalist agents.

\paragraph{Foundation Models Across Embodiments.} A central challenge in developing physical
intelligence is enabling agents to transfer
their knowledge and capabilities across
different physical embodiments.
Beyond generalization across environments and
tasks, generalist agents may need to operate
through different embodiments, including robot
arms, mobile manipulators, humanoids, and drones,
which differ in their sensing, morphology,
actuation, and action interfaces.
An important goal is to develop foundation
models whose learned world knowledge and
policies can transfer across embodiments,
rather than training separate models for
each robot.
One promising approach is to learn shared
representations of world dynamics and actions,
while using embodiment-specific components
to accommodate different observation and
action spaces. For example, Cosmos 3 uses
embodiment-specific linear projections to
map heterogeneous action representations
into a shared latent space, enabling a
common model backbone to learn from
different embodiments
~\cite{agarwal2026cosmos}.
An important question is which aspects of
world models and policies can be shared
across embodiments, and how much adaptation
is needed for new embodiments.

\paragraph{Scaling Data and Learning Signals for Generalist Agents.}
The success of foundation models suggests
that scaling data, model capacity, and
computation can be an effective path toward
generalization. For learning generalist agents,
we highlight two complementary dimensions:
\emph{source scaling}, which leverages
heterogeneous data across environments,
tasks, embodiments, and modalities, and
\emph{signal scaling}, which extracts
complementary learning signals from
the available data.
\begin{itemize}[leftmargin=*]
\item  Source scaling aims to exploit diverse
data sources, including images, passive videos,
egocentric human videos, and robot interaction
trajectories~\cite{grauman2022ego4d,o2024open}.
These sources provide complementary information:
images capture visual and semantic knowledge,
videos reveal temporal relationships and
physical dynamics, human videos capture
goal-directed interactions, and robot
trajectories additionally provide action
labels and their consequences.
Similarly, in digital environments, language,
code, browser interactions, and tool-use
trajectories provide different forms of
knowledge about the world and how agents
interact with it.
Importantly, acquiring world knowledge does
not necessarily require action-labeled data,
making it possible to leverage abundant
passive observations alongside relatively
scarce interaction trajectories.
A key challenge is how to effectively combine
these heterogeneous sources despite differences
in their representations, supervision,
and underlying environments.

\item Signal scaling aims to extract
multiple complementary learning signals
from the available data.
Beyond action prediction, interaction
trajectories can provide supervision for
future observation prediction, state-change
prediction, inverse dynamics, among others. In particular, predicting future observations
or state changes can provide much higher-dimensional
supervision than predicting relatively
low-dimensional actions.
These objectives capture different aspects
of environmental structure, dynamics,
and action consequences, potentially enabling
the model to acquire more transferable
representations.
An important question is which learning
signals are most useful, how they complement
one another, and how they should be combined
for different data sources and downstream tasks.
\end{itemize}
From this perspective, world modeling provides
an important learning signal for acquiring
knowledge about environmental dynamics,
particularly through predicting how the
environment evolves in response to actions.
However, the resulting predictive model
need not be explicitly used for planning
at deployment: the acquired knowledge may
instead support direct policy learning, or  further fine-tuning through reinforcement
learning. Thus, the value of world modeling
extends beyond its role as a predictive
simulator for planning. More broadly, these two dimensions suggest
that developing generalist agents requires
not only increasing the amount and diversity
of available data, but also improving how
effectively transferable knowledge is
extracted from that data.
Understanding how data sources and learning
signals jointly affect generalization across
environments, tasks, and embodiments remains
an important research direction.

\paragraph{Representations of States and Actions.}
An important design choice is how states and actions
are represented. Although we have distinguished
observation-space and latent-space models,
the specific representations can substantially
affect what an agent learns, how well it
generalizes, and how its predictions are
translated into executable actions.

For example, in robotics, an agent may receive
observations from cameras and proprioceptive
sensors, while its actions can be represented
in different coordinate spaces and at different
levels of abstraction. Actions may specify
low-level motor torques, joint-space targets,
or Cartesian end-effector targets, such as
desired positions and orientations, which
are converted into executable commands by
lower-level controllers~\cite{chi2025diffusion,
o2024open}. More generally, temporally extended
actions can abstract away sequences of
low-level decisions~\cite{sutton1999between}.
Recent studies have also explored using
general-purpose foundation models, such as
GPT-6 Astra and Claude, for robotic control
~\cite{yang2026astranomy,
zhang2026unexpected}, further highlighting
the importance of action representations
and control interfaces. Higher-level action
representations can simplify learning by
delegating low-level control, but may also
limit the agent's ability to exploit
fine-grained dynamics. Similarly, state representations may range
from raw observations to structured quantities,
such as object poses and joint configurations,
or learned latent representations
~\cite{finn2016deep,nair2022r3m,huang2026pointworld}.
Different representations preserve different
aspects of the environment and may vary
in their suitability for prediction, planning,
and control. Compact representations can
simplify learning but may discard information
needed for accurate dynamics prediction
or effective decision-making.

More broadly, state and action representations
determine the level of abstraction at which
an agent models and interacts with the world.
In particular, when high-level actions are
executed through lower-level controllers,
the world model learns the resulting
closed-loop dynamics rather than the direct
response to motor commands. Understanding
how these choices affect learnability,
generalization, and control is an important
direction toward developing generalist agents
across diverse environments and tasks.

\paragraph{Model-Based Planning vs  High-Level Reasoning.}
While predictive dynamic models support planning
and control, many existing approaches focus on
short-horizon prediction and action generation,
rather than high-level deliberative reasoning.
For example, DreamZero jointly predicts future
observations and actions for reactive robotic
control, while identifying long-horizon reasoning
as an important future direction~\cite{ye2026world}. Such approaches are often closer to reactive
control than to the high-level reasoning
capabilities demonstrated by LLMs,
which can leverage additional inference-time
computation to decompose complex tasks,
explore alternative strategies, and reason
about intermediate goals. Recent explorations of general-purpose reasoning
models, such as GPT-6 Astra, for robotic control
suggest a promising complementary direction
~\cite{yang2026astranomy}.
An important question is how to integrate
high-level reasoning with predictive world models,
and particularly in which representation space
such reasoning should occur: natural language,
structured physical states, learned latent
representations, or combinations thereof.
More broadly, can additional inference-time
computation improve an agent's ability to
reason about goals
and plan over longer horizons, while retaining
the efficiency required for real-time control?

\paragraph{Evaluating World Models for Decision-Making.}
Finally, a fundamental challenge is how to
evaluate the quality and usefulness of
learned world models, particularly in terms
of their contributions to decision-making.
While predictive accuracy provides a natural
evaluation criterion, it does not necessarily
reflect how well a model supports decision-making.
A world model may accurately predict observations
while failing to capture aspects of the environment
most relevant to action selection.
Conversely, a compact model that preserves
control-relevant information may support effective
policies without reconstructing the environment
in high fidelity~\cite{ferns2012metrics}. A related question is whether increasingly
high-fidelity world models, such as realistic
simulators, necessarily lead to better policies. Moreover, model errors in regions visited by
the learned policy may be more consequential
than larger errors in irrelevant regions
~\cite{janner2019trust}.
These considerations suggest that world models
should be evaluated not only by their predictive
fidelity, but also by their ability to support
planning, policy improvement, and generalization
across environments and tasks.

\section*{}

\bibliographystyle{ieeetr}
{\small
\bibliography{reference}
}

\end{document}